\documentclass{article}
\usepackage[margin=1in]{geometry}

\usepackage{natbib}
\usepackage{charter}
\usepackage[charter]{newtxmath}

\usepackage{amssymb}
\usepackage{mathrsfs}
\usepackage{multirow} 
\usepackage{graphicx}
\usepackage{booktabs}
\usepackage{rotating}
\usepackage{amsmath,amsthm}
\usepackage{xcolor}
\usepackage{colortbl}
\usepackage{hyperref}
\usepackage{url}
\usepackage{algorithm}
\usepackage[noend]{algpseudocode}
\usepackage[english]{babel}
\usepackage{amsthm}
\usepackage{tabularx}
\usepackage{cleveref}
\usepackage{graphicx}
\usepackage{caption}
\usepackage{subcaption}
\usepackage{parskip}
\definecolor{mygreen}{RGB}{76,175,80}
\definecolor{mylightgreen}{RGB}{200,230,201}
\usepackage{authblk}

\newtheorem{definition}{Definition}

\newcommand*\Let[2]{\State #1 $\gets$ #2}

\newcommand{\jrv}[0]{\alpha_{\text{JR}}}

\title{Question the Questions: \\ Auditing Representation in Online Deliberative Processes}

\author[1,2*]{Soham De}
\author[3*]{Lodewijk Gelauff}
\author[3 $\dagger$]{\\Ashish Goel}
\author[1 $\dagger$]{Smitha Milli}
\author[1,4 $\dagger$]{Ariel Procaccia}
\author[3 $\dagger$]{Alice Siu}

\affil[1]{FAIR at Meta}
\affil[2]{University of Washington}
\affil[3]{Stanford University}
\affil[4]{Harvard University}

\affil[*]{\small Equal contribution}
\affil[$\dagger$]{\small Alphabetical order}

\date{November 2025}

\begin{document}

\maketitle
\begin{abstract}

A central feature of many deliberative processes, such as citizens' assemblies and deliberative polls, is the opportunity for participants to engage directly with experts. While participants are typically invited to propose questions for expert panels, only a limited number can be selected due to time constraints. This raises the challenge of how to choose a small set of questions that best represent the interests of all participants. We introduce an auditing framework for measuring the level of representation provided by a slate of questions, based on the social choice concept known as \emph{justified representation} (JR). We present the first algorithms for auditing JR in the general utility setting, with our most efficient algorithm achieving a runtime of $O(mn\log n)$, where $n$ is the number of participants and $m$ is the number of proposed questions. We apply our auditing methods to historical deliberations, comparing the representativeness of (a) the actual questions posed to the expert panel (chosen by a moderator), (b) participants' questions chosen via integer linear programming, (c) summary questions generated by large language models (LLMs). Our results highlight both the promise and current limitations of LLMs in supporting deliberative processes. By integrating our methods into an online deliberation platform that has been used for over hundreds of deliberations across more than 50 countries, we make it easy for practitioners to audit and improve representation in future deliberations.
\end{abstract}

\includegraphics[width=\textwidth]{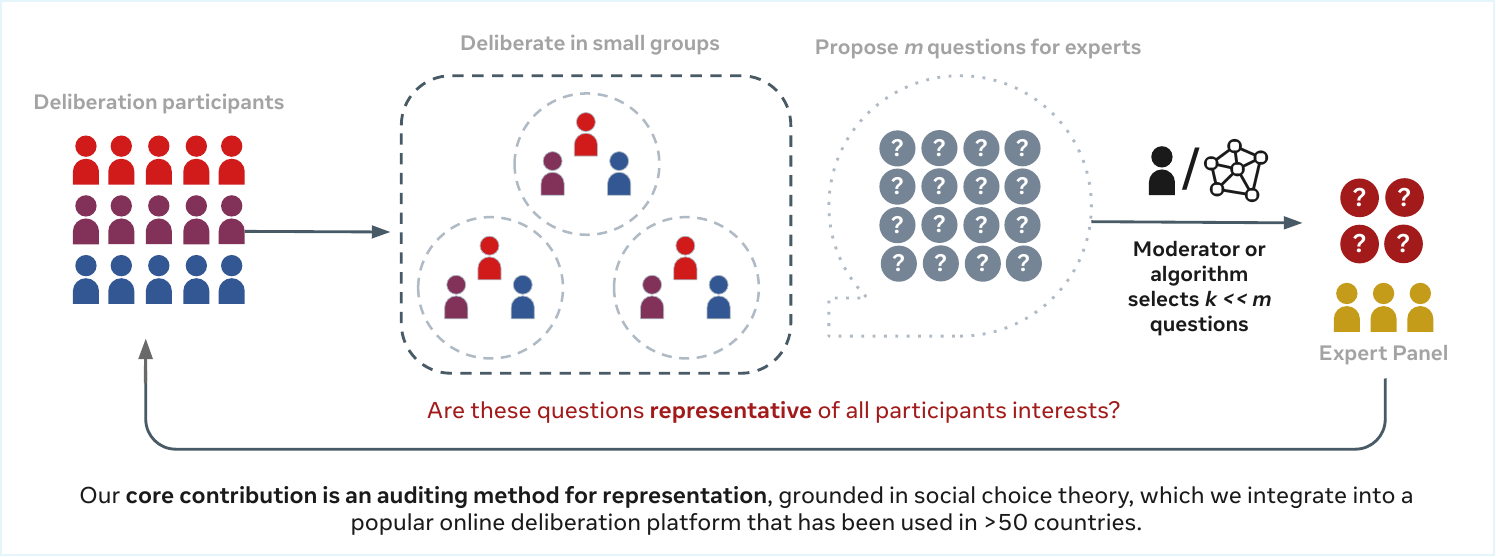}
\label{fig:teaser}

\section{Introduction}


Deliberative processes --- such as citizens' assemblies and deliberative polls~\citep{fishkin1995} --- involve randomly selected members of the public engaging in informed discussion in small groups about policy issues. These processes have gained popularity across the globe as a means to understand the reflective preferences of a population and guide collective decision-making~\citep{oecd2020innovative}. This is especially true now that technology makes it possible to organize them largely online, significantly reducing the cost of assembling a representative set of participants. 

One critical component of many deliberative designs is the opportunity for participants to interact with experts to deepen their understanding of the topic. For instance, in the deliberative polling approach, each participant (or small group) is invited to submit a question for an expert panel. However, because these events often include hundreds of participants, only a limited number of questions can ultimately be posed to the experts.

This gives rise to an important challenge: how can a small set of questions be chosen to best represent the interests and concerns of the entire participant panel? Large language models (LLMs) may offer a promising approach. In recent years, LLMs have been used to generate consensus statements in caucasus deliberations~\citep{tessler2024}, create popular summary fact-checks on Community Notes programs on social media~\citep{de2025}, and summarize public opinions on collective response systems such as Pol.is~\citep{fish2024,Jigsaw_2025,small2023opportunities}. Similarly, LLMs could be leveraged to synthesize commonalities in participant questions and generate a small slate of final questions. However, given the often nuanced and sensitive nature of deliberative processes, it is essential to rigorously audit whether such LLM-generated summary questions actually represent all participants.

\textbf{Auditing framework for representation.} We contribute an auditing framework for measuring the level of representation provided by a slate of questions. To formalize representation, we utilize a quantitative variant~\citep{bardal2025} of a widely-used concept from social choice theory known as \emph{justified representation} (JR)~\citep{aziz2017justified}. Justified representation is an axiom that requires, informally speaking, that if $k$ questions are selected, then any group of $n/k$ participants (where $n$ is the total number of participants) with shared preferences get at least one slot in the slate. We present the first algorithms for auditing JR in the general utility setting (beyond approval voting), the most efficient of which achieves an $O(mn\log n)$ runtime, where $m$ is the number of items (participant-proposed questions) and $n$ is the number of participants.

\textbf{Auditing historical deliberations.} We apply our algorithms to audit the representation of LLM-generated slates of questions in historical deliberations conducted by the Stanford Deliberative Democracy Lab. The deliberations span topics in democratic reform and generative AI. We compare the actual historical questions posed to experts (selected by a moderator) to two algorithmic approaches for question selection and generation. The first is an extractive summarization method, which selects a representative slate exclusively from participants’ original questions using an integer program that is designed to provide the optimal selection from the lens of representation (optimize the JR value). In contrast, the abstractive summarization approach synthesizes new summary questions from participant input, generating LLM-based questions that may not have been explicitly proposed.

Our findings indicate that both extractive and abstractive algorithmic methods generally yield question slates with greater representativeness than those selected by human moderators. In some cases, the abstractive approach outperforms the extractive method, while in others, the reverse holds true. This suggests that although LLM-generated summaries can sometimes enhance representativeness by synthesizing information, their benefits are not uniform. These results underscore the need for robust auditing mechanisms to determine when LLM-based methods provide added value over simply selecting from participants’ original questions --- particularly since relying on participants’ own questions may offer additional benefits in terms of trust and transparency.

\textbf{Integration into online deliberation platform.} Finally, while we evaluate on historical data, we integrate both our auditing algorithms and question selection / generation algorithms into an online deliberation platform~\citep{gelauff2023achieving} that has been used to conduct hundreds of deliberations in over 50 countries, enabling practitioners to easily utilize our approach in future deliberations.


\section{Related Work}
\textbf{AI \& deliberation.}  The success of large language models has spurred growing interest in applying AI to deliberative contexts~\citep{landemore2022can,mckinney2024integrating,small2023opportunities,korre2025evaluation}. Most relatedly, McKinney~\citep{mckinney2024integrating} proposes a framework for assessing the desirability of integrating AI tools into citizens' assemblies, focusing on two key dimensions: democratic quality and institutional capacity. In our work, we address the challenge of selecting (or generating) a set of questions for an expert panel that most effectively represents participants' interests. Rather than the development of a specific AI tool, our primary contribution is an auditing method that can be used to assess the representativeness of different approaches to question selection, whether algorithmic or human-mediated. By applying this auditing method to different algorithmic strategies, we can identify automated solutions that achieve high levels of representativeness. This focus directly supports the democratic good of inclusiveness highlighted in McKinney’s framework. Additionally, such algorithms advance McKinney's institutional goods of efficiency and scalability, as automating the question selection process eliminates the need for manual review, making it feasible to scale to a much larger number of participant questions.

As one approach to question selection, we evaluate the representativeness of LLM-generated summary questions. This is motivated by recent work using LLMs to synthesize the opinions of a collective. For example, Tessler et al. \citep{tessler2024} leverage LLMs to generate consensus statements for caucus deliberation. De et al.~\citep{de2025} build on this approach to generate summary fact-checks for Community Notes programs on social media that are preferred over human-written ones. Small et al.~\citep{small2023opportunities} discuss the opportunities and risks of integrating LLMs into pol.is~\citep{small2021}, a collective response system~\citep{ovadya2023generativecicollectiveresponse} used to understand public opinion at scale. Google Jigsaw~\citep{Jigsaw_2025} recently open-sourced a suite of LLM tools aimed at summarizing and identifying common ground in large-scale conversations such as those that take place on pol.is. Most relevant to our work, Fish et al.~\citep{fish2024} develop a method that uses LLMs to generate summary slates of opinions that satisfy certain social choice guarantees of representation.  In contrast, our focus is on developing an auditing algorithm for representation, enabling us to evaluate both algorithmic and human-mediated approaches and to quantify the potential improvements offered by algorithmic solutions.



\textbf{Social choice theory.} Our auditing framework provides a way to measure how representative a slate of questions is. To formalize representation, we draw upon the social choice concept of \emph{justified representation} (JR)~\citep{aziz2017justified}. JR is the building block that a family of subsequent axioms is based upon, e.g., PJR~\citep{sanchez2017proportional}, EJR~\citep{aziz2017justified}, EJR+~\citep{brill2023}, and BJR~\citep{fish2024}. While JR was originally introduced in the context of approval voting~\citep{aziz2017justified}, we adopt a version that accommodates general utility functions. Building on Bardal et al.~\citep{bardal2025}, we further employ a quantitative formulation of JR, which allows us to measure the degree of representativeness of a slate, rather than simply determining whether JR is satisfied. Although verifying JR in the approval voting setting is straightforward, we develop auditing algorithms tailored to the more general utility setting. 

Our focus is on auditing the JR guarantees of arbitrary slates of questions \emph{relative} to the participants’ own questions. By contrast, the work of Fish et al.~\citep{fish2024} and Boehmer et al.~\citep{boehmer2025generative} aims to generate slates that satisfy BJR over the infinite set of possible statements, but do not address the problem of auditing or verifying these guarantees. Because we seek to audit slates that may contain arbitrary questions—including those not proposed by participants, such as LLM-generated questions—we need a way to estimate participants’ utilities for these questions. To achieve this in a straightforward and interpretable manner suitable for deliberative settings, we infer a participant’s utility for a given question based on the distance between the LLM embedding of that question and the embedding of the participant’s own question. This approach bears similarity to works on proportional clustering~\citep{chen2019proportionally,micha2020proportionally,aziz2024proportionallyrepresentativeclustering}.

\section{The Auditing Framework} \label{sec:audit-framework}
Suppose there are $n$ participants in a deliberation who propose a set $Q_p \subseteq Q$ of $m$ potential questions for an expert panel, where $Q$ denotes the universe of all possible questions. Due to time constraints, however, only a small set of $k \ll m$ questions can actually be posed. The challenge, then, is to choose a set of $k$ questions $W \subseteq Q$ that best represents the interests of the overall population. In this section, we outline our framework for auditing the degree of representation that a slate of questions offers to a population of participants.  In \Cref{sec:jr}, we introduce the justified representation (JR) axiom~\citep{aziz2017justified} from social choice, along with the quantitative variant we employ~\citep{bardal2025}, which we use to formalize the notion of representation. In \Cref{sec:utility-inf}, we explain how we infer the utility that participants derive from arbitrary questions. In \Cref{sec:audit-algos}, we present the first known methods for auditing JR in the general utility setting, the best of which runs in time $O(mn \log n)$.

\subsection{Justified representation} \label{sec:jr}
To formalize what it means for a set of questions to be representative, we draw on the social choice concept of \emph{justified representation} (JR). Informally, JR is grounded in the principle of proportionality: JR requires that if $k$ questions are asked, then any group of at least $n/k$ participants---large enough to “deserve” one question by proportional allocation---who share similar preferences should have at least one question among the $k$ that represents them.

Justified representation was first introduced by Aziz et al.~\citep{aziz2017justified} in the context of approval voting, where each participant simply indicates a binary approval or disapproval for each of the $m$ candidates. However, the concept extends naturally to more general utility settings, where each participant $i$ has a utility $u_i(q) \geq 0$ for a question $q$ or has a utility $v_i(W) \geq 0$ for a slate of questions $W$. In our setting, we specifically model a participant's utility for a slate as being \emph{unit-demand}, meaning their utility is defined by their best item on the slate: $v_i(W) = \max_{q \in W} u_i(q)$.\footnote{Other options are possible, e.g., additive utilities: $v_i(W) = \sum_{q \in W} u_i(q)$. However, we observed that additive utilities tend to make JR trivially satisfied in our context, since utilities are based on cosine similarity between question embeddings (\Cref{sec:utility-inf}). As a result, even with relatively small $k$, it is unlikely that a participant’s utility for any single question would exceed their utility for even a random slate.} JR then requires that there can be no coalition of size at least $n/k$ such that the coalition's minimum utility for an alternative question is greater than the coalition's maximum utility for the given slate. The formal definition is as follows:\\

\begin{definition}[JR with utilities]
\label{def:JR}
A slate of questions $W \subseteq Q$ satisfies JR if for every coalition $S$ of size at least $n/k$, there is no alternative question $q \in Q_p$ such that $\min_{i\in S} u_i(q)> \max_{i\in S} v_i(W)$. \label{def:jr}
\end{definition}
\Cref{def:jr} departs slightly from the standard definition of JR by allowing the slate $W$ to consist of arbitrary questions from the entire space $Q$, rather than being restricted to the discrete set $Q_p$ of participant-proposed questions.\footnote{In this case, satisfying JR means that there is no sufficiently large, cohesive coalition that would prefer to deviate to a participant-proposed question in $Q_p$. However, it does not guarantee that there is no coalition that prefers some other question in the infinite set $Q$ of all potential questions~\citep{fish2024,boehmer2025generative}.} In our setting, we will also audit slates of LLM-generated summary questions that synthesize, but are distinct from, the participants' original questions.

As originally proposed, JR provides only a binary notion of whether a slate is representative or not. However, we would also like to quantify the degree of representation offered by different question selection methods. Following Bardal et al.~\cite{bardal2025}, we quantify JR by identifying the smallest size group for which a JR guarantee holds. Specifically, we find the smallest size $\alpha$ for which $\alpha$-JR is satisfied, where $\alpha$-JR generalizes JR by providing guarantees to coalitions of size at least $\alpha \cdot n/ k$. When $\alpha = 1$, we recover the original JR definition, guaranteeing representation only to groups of size at least $n/k$. When $\alpha < 1$, the guarantee extends to smaller-sized coalitions. \\

\begin{definition}[$\alpha$-JR with utilities]
\label{def:JR_beta}
    A slate of questions $W \subseteq Q$ satisfies $\alpha$-JR if for every coalition $S$ of size at least $\alpha \cdot n/ k$ there is no alternative question $q \in Q_p$ such that $\min_{i\in S} u_i(q)> \max_{i\in S} v_i(W)$.
\end{definition}

We audit the level of representation that a slate $W$ provides by calculating the highest value of $\alpha$ for which it satisfies $\alpha$-JR. We refer to this optimal value as the \emph{JR value of $W$}: $\jrv(W) = \inf \{\alpha : W \text{ satisfies } \alpha\text{-JR} \}$.

\subsection{Utility inferences} \label{sec:utility-inf}
In order to measure the JR value of an arbitrary slate $W \subseteq Q$, we must be able to infer the utility that a participant has for any potential question $q \in Q$, beyond just the participants' own proposed questions. While complex models for predicting user utility are common in domains like recommender systems~\citep{zhang2019}, in the context of deliberation, there is usually very limited information available about participants to support such inferences. Furthermore, transparency is crucial; it must be possible to clearly explain how utility estimates are derived to both moderators and participants~\citep{mckinney2024integrating}.

Given these constraints, we adopt a simple and interpretable approach: we infer a participant’s utility for a question based on its similarity to their own proposed question. Specifically, each participant $i$ is associated with their proposed question $q_i$, and we then measure the participant's utility $u_i(q')$ for an alternate question $q'$ is defined as $u_i(q') = s(q_i, q')$ where $s$ is a similarity measure. In our implementation, we use cosine similarity between LLM-generated embeddings of the two questions.

\begin{figure}[h]
    \centering
    \includegraphics[height=6cm]{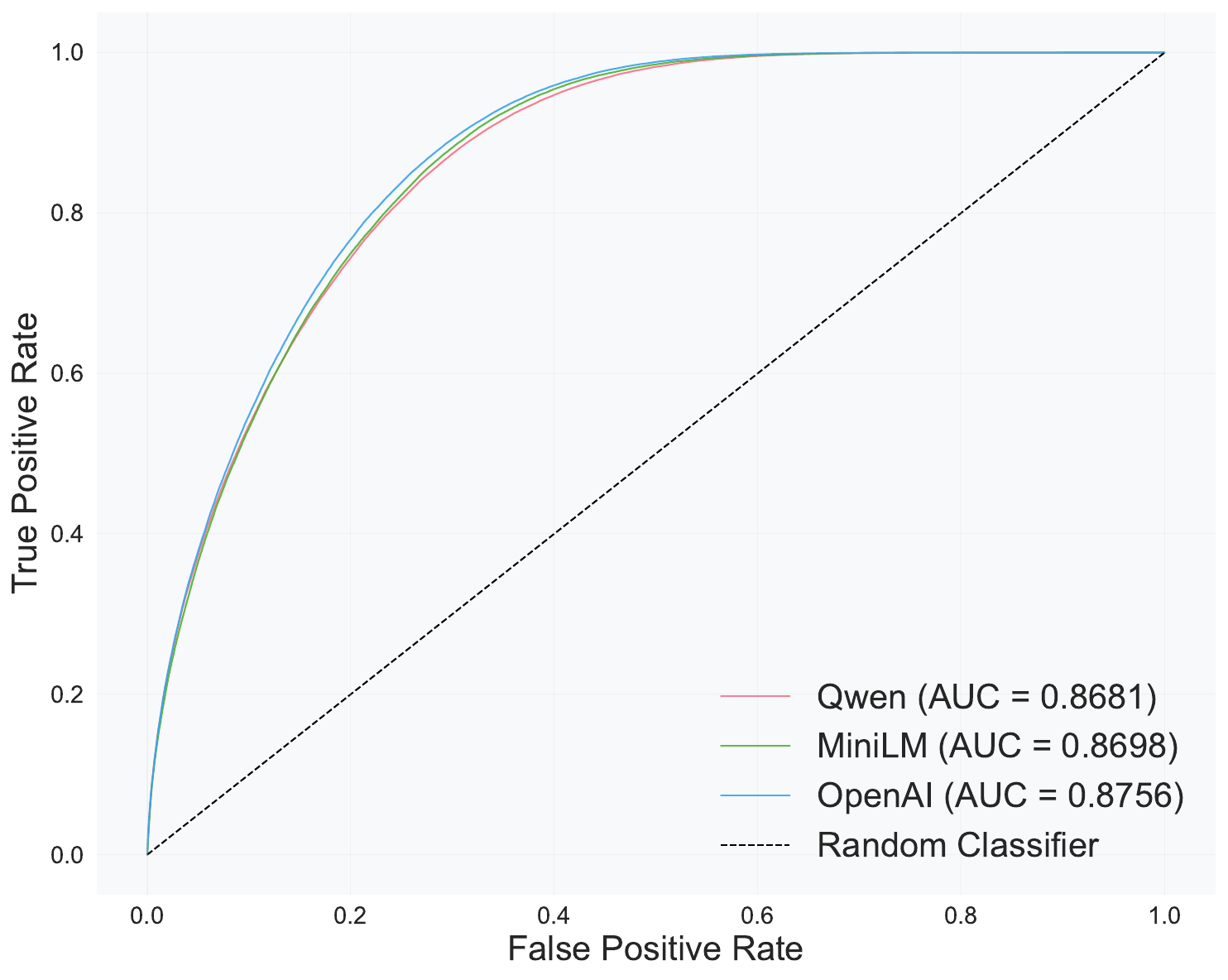}
    \caption{ROC curves comparing the binary classification accuracy of different embedding models on the Quora Question Pairs (QQP) dataset \cite{quora_question_pairs}.  Each curve is obtained by thresholding the cosine similarity between the embeddings of paired questions.}
    \label{fig:qqp_eval}
\end{figure}

\begin{figure*}[t]
    \centering
    \includegraphics[width=\textwidth]{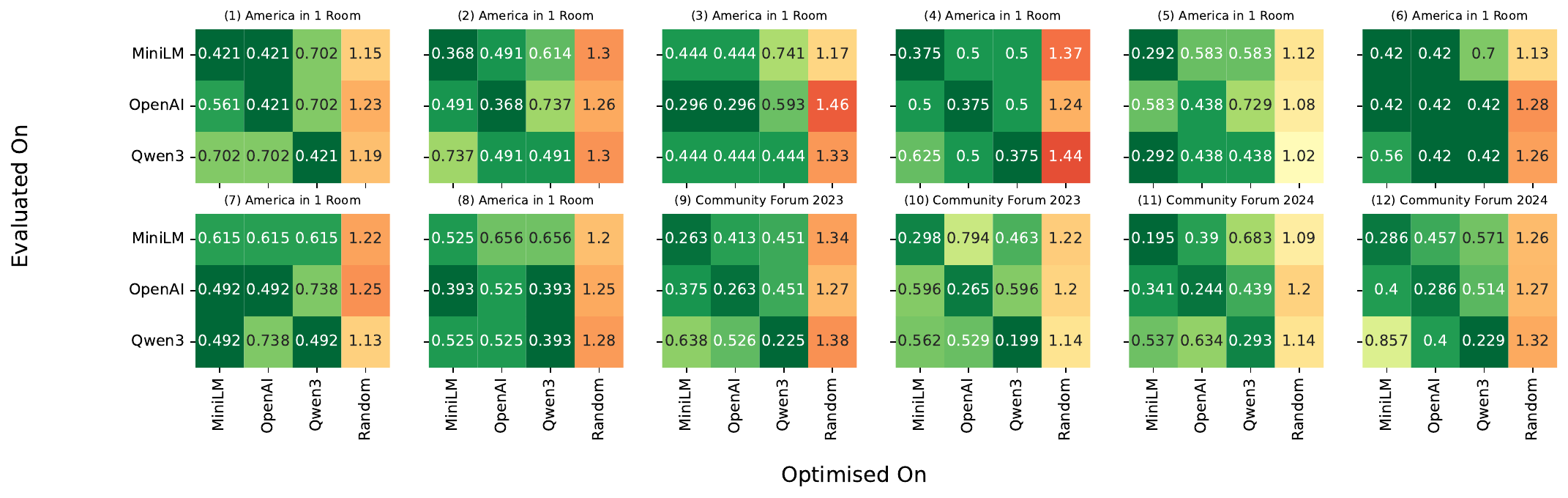}
    \caption{Cross-validating audit outcomes across embedding models. Each heatmap shows the JR-value when slates optimized using an integer program on one embedding model are evaluated using another model.  Rows indicate the model used for evaluation, and columns indicate the model used for optimization (via the IP). Lower JR-values indicate greater consistency between optimization and evaluation models. Any value below 1 implies that the slate satisfies JR.}
    \label{fig:cross_val_oai}
\end{figure*}

Cosine similarity between modern sentence embeddings has been shown to correlate well with human similarity judgments~\citep{gao-etal-2021-simcse}. To validate our approach in the question context, we evaluate how effectively cosine similarity can differentiate between duplicate and non-duplicate questions using the Quora question pairs dataset, which includes human-annotated binary labels (0 for different meanings, 1 for the same meaning)\cite{quora_question_pairs}. Figure~\ref{fig:qqp_eval} presents the ROC curves generated by varying the cosine similarity threshold across three different embedding models (Qwen3 0.6B \cite{zhang2025qwen3}, all-MiniLM-L6-v2 and text-embedding-3-small (OpenAI)). All embedding models achieve high AUC scores between 0.868 and 0.876. To demonstrate that our auditing approach (see \Cref{sec:audit-algos}) yields consistent outcomes across different embedding models, we apply it  to slates identified as most representative by any given embedding model (using the integer program described in \Cref{appendix:integer-program}). We find that these slates yield similarly low JR-values when evaluated with other widely-used embedding models (\Cref{fig:cross_val_oai}).

\subsection{Auditing algorithms} \label{sec:audit-algos}
We now introduce our algorithms for auditing the JR value of a slate. To our knowledge, these algorithms are also the first algorithms for verifying JR in the general utility setting. For binary (approval) utilities, verifying whether a slate $W$ satisfies JR is straightforward~\citep{aziz2017justified}: one simply checks whether there exists an alternative $q \in Q_p$ that is approved by at least $n/k$ participants, all of whom disapprove every question in the slate. In contrast, for general additive utilities, the verification problem is more complex. A JR violation occurs when there exists an alternative $q \in Q_p$, a utility threshold $\tau > 0$, and a group of participants $S$ of size at least $n/k$ where, for every participant $i \in S$, their utility for the alternative $u_i(q)$ is greater than $\tau$, while their utility for the slate $v_i(W)$ is at most $\tau$.

\textbf{Naive algorithm.} At first glance, it may seem necessary to check an infinite set of possible utility thresholds $\tau$. However, it suffices to consider only the thresholds defined by the participants' utilities for the slate, i.e., the set of $n$ thresholds: $T = \{v_i(W)\}_{i \in [n]}$. For each alternative $q \in Q_p$, we can iterate over each threshold $\tau \in T$ to identify the number of participants whose utility for the alternative $q$ exceeds $\tau$, while their utility for the slate $W$ is at most $\tau$. While doing so, we keep track of the largest size of a deviating coalition, which allows us to find the JR value of the slate. This approach yields the simple $O(mn^2)$ algorithm described in \Cref{alg:verify-fjr-1}.

\begin{algorithm}
  \caption{Compute $\jrv(W)$ in $O(mn^2)$ time}
  \label{alg:verify-fjr-1}
  \begin{algorithmic}[1]
    \Require Slate $W$, participant questions $Q_p$, utility function $u_i : Q \rightarrow \mathbb{R}_{\geq 0}$ for each participant $i \in [n]$
    \Let{$\alpha$}{$0$}
    \ForAll{$q \in Q_p$}
      \For{$i = 1$ to $n$}
        \Let{$\tau$}{$v_i(W)$}
        \Let{$c$}{$|\{j \in [n] : u_j(q) > \tau \text{ and }  v_j(W) \leq \tau\}|$}
        \Let{$\alpha$}{$\max(\alpha, c \cdot k / n)$}
      \EndFor
    \EndFor
    \Return $\alpha$
  \end{algorithmic}
\end{algorithm}


\textbf{Single-pass algorithm.} We now propose an improved algorithm that achieves an $O(mn\log n)$ runtime (Algorithm~\ref{alg:verify-fjr-2}). The previous approach (Algorithm~\ref{alg:verify-fjr-1}) iterates over each alternative question $q \in Q_p$ and, for each possible utility threshold $\tau \in T$, determines the largest coalition of participants who would prefer $q$ to $W$. This results in a higher computational cost, as each of the $n$ thresholds is considered separately. In contrast, Algorithm~\ref{alg:verify-fjr-2} exploits the observation that, for each question $q$, the largest blocking coalition can be identified in a single pass over the sorted utility thresholds. 

First, the participants are sorted in non-increasing order according to their utility for the given slate $W$, resulting in the ordering $\gamma_1, \dots, \gamma_n \in [n]$ where $v_{\gamma_1}(W) \geq \cdots \geq v_{\gamma_n}(W)$. For each alternative question $q \in Q_p$, the participants are also sorted in non-increasing order of their utility for $q$, resulting in the ordering $\delta_1, \dots, \delta_n \in [n]$ where $u_{\delta_1}(q) \geq \cdots \geq u_{\delta_n}(q)$. The algorithm then iterates through the sorted utilities $\{u_{\delta_i}(q)\}_{i=1}^n$, maintaining a counter $c_q$ for the size of the largest coalition that prefers the question $q$ to the slate $W$. Here, rather than checking each utility threshold $\{v_{\gamma_t}(W)\}_{t=1}^n$ separately, we iterate over them in one pass, starting with $t = 1$. Each participant's utility $u_{\delta_i}(q)$ for $q$ is compared to the current threshold $v_{\gamma_t}(W)$. If $u_{\delta_i}(q) > v_{\gamma_t}(W)$, then the participant is added to the coalition, and the counter is incremented. If $u_{\delta_i}(q) \leq v_{\gamma_t}(W)$, then this means that the coalition size cannot be increased without lowering the threshold. Thus, participant $\gamma_t$ is blacklisted from future coalitions (and removed from the existing coalition, if applicable), and the utility threshold is decreased by incrementing $t$.

This process continues until all participants have been considered or all thresholds have been exhausted. The maximum coalition size $c$ encountered during this process for any question $q$ is recorded, and the algorithm returns $\jrv(W) = \frac{c}{n/k}$. By leveraging sorted orderings and a single-pass approach, this algorithm achieves a significant improvement in efficiency, enabling the exhaustive verification of JR for a large number of candidate slates.

\begin{algorithm}
  \caption{Compute $\jrv(W)$ in $O(mn\log n)$ time}
  \label{alg:verify-fjr-2}
  \begin{algorithmic}[1]
    \Require Slate $W$, participant questions $Q_p$, utility function $u_i : Q \rightarrow \mathbb{R}_{\geq 0}$ for each participant $i \in [n]$
    \Let{$\gamma_1, \dots, \gamma_n$}{Participants sorted such that $v_{\gamma_1}(W) \geq \cdots \geq v_{\gamma_n}(W)$}
    \Let{$c$}{$0$}
    \ForAll{$q \in Q_p$}
      \Let{$\delta_1, \dots, \delta_n$}{Participants sorted s.t. $u_{\delta_1}(q) \geq \cdots \geq u_{\delta_n}(q)$}
      \Let{$c_q$}{0} \Comment{Size of deviating coalition for question $q$}
      \Let{$S, B$}{$\emptyset, \emptyset$} \Comment{Current coalition and blacklisted participants}
      \Let{$t$}{$0$}
      \For{$i = 1$ to $n$}
        \While{$t < n$ and $u_i(x) \leq v_{\gamma_t}(W)$} 
          \If{$\gamma_t \in S$}
            \State $c_q \leftarrow c_q - 1$
          \EndIf
          \State $B \leftarrow B \cup \{ \gamma_t \}$
          \State $t \leftarrow t + 1$
        \EndWhile
        \If{$t \geq n$}
            \State Break
        \EndIf
        \If{$i \notin B$}
            \State $S \leftarrow S \cup \{i\}$
            \State $c_q \leftarrow c_q + 1$
            \State $c \leftarrow \max(c_q, c)$
        \EndIf
        
      \EndFor
    \EndFor
    \Return $c \cdot k/n$
  \end{algorithmic}
\end{algorithm}

\section{Empirical Evaluation}
\label{sec:evals}
In the following sections, we first describe the platform and deliberative polls (\Cref{sec:platform-integration}) that provide the data used for our audit. Then, we introduce the relevant baselines and present the key results from applying our audit to this data (\Cref{sec:evals-summaries}).

\subsection{Deliberative Polls and the Online Deliberation Platform}
\label{sec:platform-integration}
\begin{figure*}[!h]
    \centering
    \includegraphics[width=\textwidth]{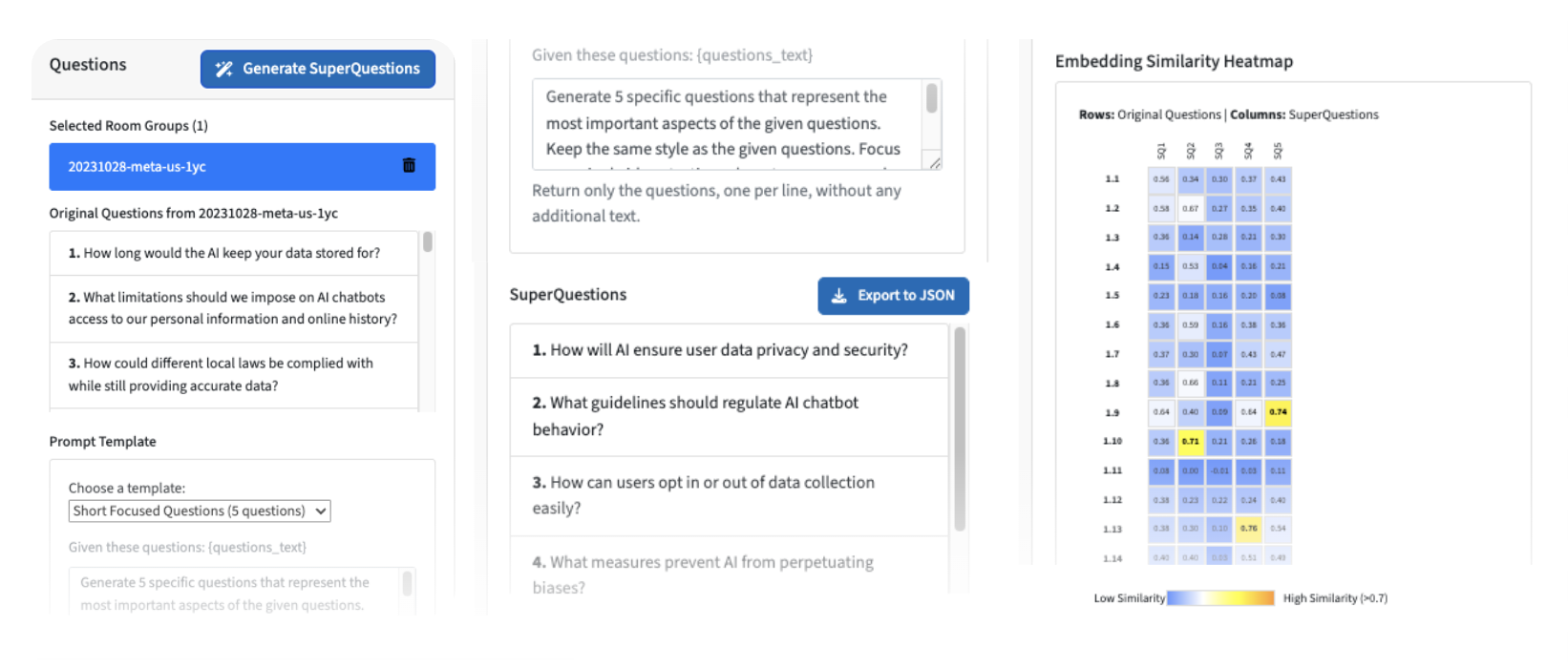}
    \caption{Screenshots illustrating our approach implemented in the online deliberation platform. The moderator can generate LLM summary questions (referred to as ``SuperQuestions'' in the interface) from participant-proposed questions; view which participant-proposed questions are most similar to each LLM-generated one; and export all data including, similarity scores, for representation auditing.}
    \label{fig:platform_screenshot}
\end{figure*}
We evaluate our audit on data from 12 sessions in 3 different \emph{deliberative polls}~\citep{fishkin1995,luskin2002}, conducted by the Stanford Deliberative Democracy Lab, that cover a wide range of topics — from the American political system to AI agents. During such a deliberative poll, participants are assigned to small groups with on average 10 participants; each group has a conversation, at the end of which they propose and rank questions for a plenary panel of experts. The plenary panel then is presented a slate of 5--15 of these questions that helps clarify the topic for the participants. Moderators may also occasionally group multiple very-similar `sibling' \footnote{In human-generated slates with a set of  sibling questions, we consider all possible slates that can be formed by taking one sibling question as the representative of its sibling set and report the average JR value for the Human baseline in \Cref{tab:oai_results_gurobi_alpha}.} questions together and pose them to the panel. It is this set of selected questions that we use to evaluate our algorithms. In \Cref{sec:evals-summaries}, we evaluate the JR value of the slates of question selected by moderators, compared to the slates of questions generated by algorithmic approaches.

We consider datasets based on sessions from 3 deliberations: ``America in One Room: Democratic Reform'' (June 2023), the ``Meta Community Forum on AI chatbots'' (two sessions from October 2023), and the ``Meta Community Forum on AI agents'' (two sessions from October 2024).  
\begin{itemize}
    \item The America in One Room (A1R) deliberations convened U.S. nationally representative sample of participants to discuss potential reforms to the American political system, such as changes to the electoral college and the adoption of ranked-choice voting~\citep{Fishkin_Diamond_2023}. This results in 8 sets of questions and corresponding slates.
    \item  The 2023 Meta Community Forum (CF'23) convened nationally representative samples of participants from Brazil, Germany, Spain and the United States to discuss policies for user interactions of AI chatbots in their respective national languages in small groups. In two sessions of small groups, the participants proposed questions, followed by an international plenary panel of experts~\citep{chang2024}. This results in 2 sets of questions and corresponding slates.
    \item The 2024 Meta Community Forum (CF'24) similarly convened a nationally representative sample of participants from Brazil, India, Nigeria, Saudi Arabia, and South Africa that discussed policies and conditions for AI agents~\citep{chang2025}. This results in 2 sets of questions and corresponding slates.
\end{itemize}

All deliberations were conducted on an online, self-moderating deliberation platform that has been shown to perform on par with human moderators across multiple dimensions~\citep{gelauff2023achieving}. We implemented our auditing algorithms, as well as algorithmic summarization using LLMs, within this same online deliberation platform (Figure~\ref{fig:platform_screenshot}). The goal of the implementation is to provide an easy way for moderators to generate and evaluate different potential slate of questions to pose to an expert panel. For each generated slate, the platform computes the JR value $\jrv(W)$ and displays a heatmap illustrating the similarity between each selected question and those submitted by participants. This feature enables moderators, when presenting a question, to point out which participants submitted similar questions and to explicitly highlight how participants' own contributions informed the final questions that were posed.


\begin{table}[ht]
    \centering
    \setlength{\tabcolsep}{2pt}
    \begin{tabular}{@{}l|ccccc|c|c@{}}
        \toprule 
        \textbf{Panel} & \textbf{Human} & \textbf{Random} & \textbf{IP} & \textbf{LLM} & \textbf{LLM$_{\text{best}}$} & \textbf{N} & \textbf{k} \\
        \midrule
        A1R & 0.842 & 1.227 \text{\scriptsize{$\pm$ 0.077}} & \cellcolor{mygreen}0.421 & 0.778 \text{\scriptsize{$\pm$ 0.032}} & \cellcolor{mygreen}0.421 & 57 & 8\\ 
        A1R & 0.798$\dag$ & 1.258 \text{\scriptsize{$\pm$ 0.083}} & \cellcolor{mygreen}0.368 & 0.57 \text{\scriptsize{$\pm$ 0.023}} & \cellcolor{mygreen}0.368 & 57 & 7\\ 
        A1R & 1.037 & 1.459 \text{\scriptsize{$\pm$ 0.099}} & \cellcolor{mygreen}0.296 & 0.913 \text{\scriptsize{$\pm$ 0.052}} & \cellcolor{mylightgreen}0.444 & 54 & 8\\ 
        A1R & 0.875$\dag$ & 1.241 \text{\scriptsize{$\pm$ 0.096}} & \cellcolor{mygreen}0.375 & 0.701 \text{\scriptsize{$\pm$ 0.039}} & \cellcolor{mygreen}0.375 & 56 & 7\\ 
        A1R & 1.021 & 1.084 \text{\scriptsize{$\pm$ 0.064}} & \cellcolor{mygreen}0.438 & 1.018 \text{\scriptsize{$\pm$ 0.044}} & \cellcolor{mylightgreen}0.583 & 48 & 7\\ 
        A1R & 1.68$\dag$ & 1.281 \text{\scriptsize{$\pm$ 0.091}} & \cellcolor{mylightgreen}0.42 & 0.545 \text{\scriptsize{$\pm$ 0.022}} & \cellcolor{mygreen}0.28 & 50 & 7\\ 
        A1R & 0.738 & 1.247 \text{\scriptsize{$\pm$ 0.076}} & \cellcolor{mylightgreen}0.492 & 0.751 \text{\scriptsize{$\pm$ 0.041}} & \cellcolor{mygreen}0.369 & 65 & 8\\ 
        A1R & 1.049 & 1.246 \text{\scriptsize{$\pm$ 0.079}} & \cellcolor{mylightgreen}0.525 & 0.744 \text{\scriptsize{$\pm$ 0.032}} & \cellcolor{mygreen}0.393 & 61 & 8\\ 
        CF'23 & 2.234$\dag$ & 1.267 \text{\scriptsize{$\pm$ 0.076}} & \cellcolor{mygreen}0.263 & 1.052 \text{\scriptsize{$\pm$ 0.051}} & \cellcolor{mylightgreen}0.563 & 293 & 11\\ 
        CF'23 & 0.827$\dag$ & 1.197 \text{\scriptsize{$\pm$ 0.075}} & \cellcolor{mygreen}0.265 & 1.05 \text{\scriptsize{$\pm$ 0.043}} & \cellcolor{mylightgreen}0.562 & 272 & 9\\ 
        CF'24 & 1.268 & 1.203 \text{\scriptsize{$\pm$ 0.067}} & \cellcolor{mygreen}0.244 & 0.948 \text{\scriptsize{$\pm$ 0.042}} & \cellcolor{mylightgreen}0.537 & 164 & 8\\ 
        CF'24 & 0.8 & 1.267 \text{\scriptsize{$\pm$ 0.074}} & \cellcolor{mygreen}0.286 & 0.785 \text{\scriptsize{$\pm$ 0.056}} & \cellcolor{mylightgreen}0.4 & 175 & 10\\ 
                \bottomrule
    \end{tabular}

\caption{Levels of representation ($\jrv$) for different slates using embeddings from the \texttt{text-embedding-3-small} model by OpenAI. 
Baselines (Random and LLM-generated slates) reflect mean JR values over 100 runs, with 95\% confidence intervals. In sessions marked with $\dag$, human moderators posed multiple \textit{sibling} questions.}
\label{tab:oai_results_gurobi_alpha}
\vspace{0.5em}
\end{table}

\subsection{Evaluating algorithms for question selection and generation}
\label{sec:evals-summaries}
A representative slate of $k$ questions that summarizes all the $n$ questions proposed by participants in a deliberation can emerge via two distinct pathways. First, we may select a representative subset of $k$ questions from the larger set of all proposed questions. We call the resulting slate an \emph{extractive summarization} as it is formed by directly extracting from the set of all questions proposed by participants. Alternatively, we can generate $k$ new questions that aim to summarize and highlight the important aspects of proposed questions but may not exist verbatim in the original set of $n$ questions. Since these questions are formed via synthesis or abstraction, we call this slate an \emph{abstractive summarization}.  

We test the slates generated through five methods, the first three of which are extractive summarization methods, and the last two of which are abstractive.
\begin{enumerate}
    \item \emph{Random.} A baseline that picks from participant-proposed questions uniformly at random
    \item \emph{Human.} The historical questions posed to the expert panel, that were selected by the moderator.
    \item \emph{IP.}  Following Theorem 2 in Bardal et al. \cite{bardal2025}, we observe that it is NP-hard to determine existence of a slate satisfying $\alpha\text{-JR}$ for a given $\alpha$. Hence, instead of a polynomial-time algorithm we implement a polynomial-size integer program, as described in \Cref{appendix:integer-program}, to determine the subset of participant-proposed questions that minimizes the JR value. In \Cref{tab:oai_results_gurobi_alpha}, we run our integer program using Gurobi and with all possible thresholds discretized at intervals of 0.01.
    \item \emph{LLM.} We prompt \texttt{gpt-4o} (see Appendix~\ref{appendix:gpt-prompts}) with all participant-proposed questions to generate $k$ representative questions. We sample multiple times while shuffling the order of questions with a temperature of one, resulting in 100 generated slates.
    \item \emph{LLM$_{best}$.} We pick the LLM-generated slate with the highest JR value, i.e., best-of-n sampling.
\end{enumerate}


\begin{table*}[!ht]
\small
\centering
\begin{tabular}{p{.8cm}p{.5cm}p{14cm}}
\toprule

\textbf{Type} & $\jrv$ & \textbf{Proposed Question Slates (A1R - Session 1)} \\
\midrule
\multirow{8}{.5cm}{\centering IP} & \multirow{8}{.5cm}{\centering 0.421} & 1. Rank choice voting seems like a good idea. Can you please discuss two pros and two cons to RCV. \\
& & 2. Is there anything that prevents using RCV (Ranked Choice Voting) in party primaries? IE Federal Law, Funding Issues, Party Bylaws? \\
& & 3. How can we change the system to benefit all Americans in their voting right? \\
& & 4. Do you think proportional representation would help mitigate the problem of Gerrymandering? \\
& & 5. Out of each of the proposals, which is considered the most cost-effective way to improve voting? \\
& & 6. What is the best way for a third party (or multiple parties) to gain a foothold in the current two-party system? Which of these proposals can make that happen? \\
& & 7. What does the political science research tell us about the difficulty of teaching voters how to operate in new systems, such as RCV, proportional representation, or changing to non-partisan primaries? How is Proportional Representation counted? \\
& & 8. How successful is rcv in places where it has been used? \\
\midrule
\multirow{8}{.5cm}{\centering Human} & \multirow{8}{.5cm}{\centering 0.842} & 1. What is the approximate cost to update voting machines to RCV? \\
& & 2. what are examples of where rank choice voting has implemented and how well do people like it? \\
& & 3. How do you best balance an rcv election where some rank and some dont rank and is that a fair and equal vote? \\
& & 4. How can we trust the PR system to actually vote for the people? How will the people be educated on how it works? \\
& & 5. In a proportional representation system at the congressional level, what would be some examples of implementation? In a multi- member district with 3 members, would all three members confer before voting as one block, or all vote with 1/3 of a vote? \\
& & 6. The group in general seems highly against Gerrymandering. What can the experts tell us about the effects of any of these proposals to eliminate or reduce gerrymandering? \\
& & 7. What is the best way for a third party (or multiple parties) to gain a foothold in the current two-party system? Which of these proposals can make that happen? \\
& & 8. How soon would any changes to elections be implemented? Is this something we would do for the next Presidential election or are we 5-10 years away from these changes going into effect? \\
\midrule
\multirow{8}{.5cm}{\centering LLM$_\text{best}$} & \multirow{8}{.5cm}{\centering 0.421} & 1. What are the most cost-effective ways to implement Ranked Choice Voting? \\
& & 2. How can we address voter education challenges for new systems like RCV and proportional representation? \\
& & 3. Which electoral reforms could most effectively reduce gerrymandering and partisanship? \\
& & 4. How successful has RCV been in areas where it has been implemented, and what feedback has it received? \\
& & 5. How would proportional representation impact party systems and mitigate political polarization? \\
& & 6. What steps are needed to ensure fair enforcement of new voting systems and prevent political corruption? \\
& & 7. Would implementing RCV require significant updates to voting machines, and what are the costs? \\
& & 8. How can third-party and independent candidates gain traction in a two-party-dominated system? \\
\bottomrule
\end{tabular}
\caption{Questions in extractive and abstractive slates from the first panel in the A1R deliberation, as evaluated in \Cref{tab:oai_results_gurobi_alpha}}
\label{tab:example_questions}
\vspace{0.5em}
\end{table*}

 Table~\ref{tab:oai_results_gurobi_alpha} presents results of running our audit on different question slates generated on data from 8 panels during the ``America in One Room: Democratic Reform'' (A1R) deliberation, 2 panels during the ``2023 Meta Community Forum on AI Chatbots'' (CF'23) deliberation and 2 panels during the ``2024 Meta Community Forum on AI Agents'' (CF'24) deliberation. 
 
\paragraph{$\triangleright$ Historical performance by human moderators} These results reveal that slates selected by human moderators are, on average, more representative than a set of $k$ randomly selected questions with a few key exceptions (such as in the first Meta'23 deliberation). However, we also note that in half of the panels on which we run our audit, human-moderator-selected slates do not satisfy JR, indicating a potential to improve the deliberative polling process by using more representative slates.

\paragraph{$\triangleright$ Abstractive vs.~extractive slates} The slates selected by the integer program are the most representative extractive slate, which by definition must always have a $\jrv \leq 1$. Similarly, the LLM$_{\text{best}}$ slates are the most representative abstractive slates out of 100 LLM-generated candidates --- an approximation of the most representative abstractive slate. We find that all algorithmic approaches to slate generation result in more representative slates when compared to random and human-selected slates that were historically used in the deliberative polls. Moreover, we find that abstractive slates are often able to match or surpass the representativeness of the best extractive slates, especially for sessions with fewer questions proposed. We also note that in sessions with more participants, abstractive slates are much more feasible to generate in real time using LLMs, whereas finding the best extractive slate might be computationally intractable at interactive speeds. Relatedly, as the example in \Cref{tab:example_questions} reveals, abstractive question slates generated using LLMs typically appear more stylistically coherent and well composed compared to extractive slates, even when they achieve equivalent levels of representativeness.

Our results highlight the promise and limitations of using LLMs for generating slates in large online deliberative process.  While LLM-generated abstractive slates consistently outperform existing human-driven processes, they do not uniformly surpass extractive slates selected by an integer program across all panels. This variability underscores the value of our efficient auditing algorithm, which enables a hybrid approach that leverages the strengths of both abstractive and extractive methods.

\section{Conclusion}
In this work, we address the challenge of selecting questions that are representative of all participants' interests in large-scale deliberative processes. We present two complementary approaches towards solving this task --- extractive (selecting from existing questions) and abstractive (generating new questions) --- and develop a framework to evaluate their representativeness. To compare these approaches, we introduce an auditing framework grounded in the social choice concept of justified representation (JR). We present the first computationally efficient algorithms for auditing JR in general utility settings. Our framework uses widely available text-embedding models to infer participant utilities based on semantic similarity, enabling scalable application across diverse deliberative contexts. This semantics-based approach provides interpretable utility measurements and avoids potential biases that may arise from black-box predictive models. Our framework ensures that every participant who proposes a question during a deliberation is appropriately represented by the final slate of questions proposed to expert panelists.

Applying our methods to several historical deliberations retroactively, we demonstrate that algorithmically selected or generated questions have the potential to enhance representativeness, compared to those chosen by human moderators. This also highlights the utility of our auditing approach, which enables moderators to compare several algorithmically generated slates and evaluate them for representativeness. Finally, we integrate our framework into a widely-used online deliberation platform that has supported hundreds of deliberations in over 50 countries, making our methods readily accessible to improve representation in future deliberations.

\textbf{Limitations and future work.}  Our work is not without limitations and we identify several promising directions for future research. First, our evaluation in \Cref{sec:evals} considers the JR value for LLM-generated and IP-selected questions separately; future work could explore hybrid approaches that select from both participant-proposed and LLM-generated questions. Second, we infer a participant's utility for a question using the cosine similarity between the embeddings of that question and the participant's proposed question. We note that our retrospective analysis of historical slates limits our ability to validate utility inferences with data from actual participants involved in the deliberation. While we evaluate our embedding-based approach on existing datasets with human similarity judgments (see \Cref{fig:qqp_eval}) and across different embedding models (\Cref{fig:cross_val_oai}), future studies could directly validate these utility inferences with participants themselves during an ongoing deliberation. Finally, although our focus is on auditing the JR value, future work could explore auditing other axioms offering stronger representation guarantees such as 
BJR~\citep{fish2024,boehmer2025generative}.

\section{Acknowledgment}

 The authors would like to thank Harshvardhan Agarwal, Zixin Xu, and the other developers for their contributions to the Stanford Online Deliberation Platform, which was instrumental in data collection. Special thanks are extended to Harshvardhan Agarwal for implementing the algorithm. The authors also acknowledge Dhruv Gupta and Advay Ranade for their assistance with data collection for the human benchmarks. Finally, the authors are grateful to Audrey Tang for encouraging the exploration of this research direction. This work was supported in part by NSF Award \#2333849.
 
\clearpage
\newpage
\bibliographystyle{plainnat}
\bibliography{refs}

\clearpage
\newpage
\section{Prompts}
\label{appendix:gpt-prompts}


Below is the prompt we use for generating LLM summary questions.
\begin{verbatim}
Given these questions : \n
{questions} \n
Generate {k} specific concise questions that exhaustively summarize these given questions 
as much as possible. Avoid long and generic high-level questions. Retain the same style, 
specificity  and length as the given questions. Return only the questions, one per line, 
without any additional text.
\end{verbatim}

\subsection{Integer Program: Minimizing Largest Dissatisfied Coherent Set}
\label{appendix:integer-program}
Using the standard notation in social choice, we use the term ``candidates'' to refer to potential questions, and the term ``voter'' to refer to each question in the original data set. In our case, the set of candidates and voters is the same. We use similarity as the notion of utility, and assume that it is given to us (e.g. via embeddings). We will use $N$ to refer to the number of voters, and $M$ to the number of candidates for generality, even though these are the same for our use case.

The integer program is modeled after a similar integer program~\cite{jiang-personal} used to devise approximation algorithms for justified representation problems. We use binary integer variables to indicate whether a particular candidate is chosen. The central idea is to use auxiliary binary integer variables to keep track, for each utility level, whether a voter has at least that much utility given the set of chosen questions. This allows us to test, for each potential blocking question that is not chosen, whether that question would result in a blocking set of voters, that is, a set of voters whose minimum utility for the blocking question exceeds what all of them are currently receiving. We minimize the size of the largest blocking set, which allows us to optimize for the over-representation factor $\beta$ that we are auditing for.

\paragraph{Given}\ 
\begin{itemize}
  \item $K$: Committee size
  \item $u(v,c)$: Utility of voter $v$ from candidate $c$ (cosine similarity)
\end{itemize}

\paragraph{Indices used}\ 
\begin{itemize}
  \item $c$: Candidates
  \item $v$: Voters
  \item $s$: Satisfaction level, ranges over all possible satisfaction levels (at most $NM$)
  \item $c'$: Blocking candidate
\end{itemize}

\paragraph{Variables}\ 
\begin{itemize}
  \item $J$: Size of largest dissatisfied coherent set; integer variable
  \item $x_c$: Indicator variable, 1 if candidate $c$ is in the committee; binary integer variable
  \item $y_{v,s}$: Indicator variable, 1 if voter $v$ has utility at least $s$; binary integer variable
\end{itemize}

\subsubsection*{Integer Program}
\begin{align*}
\text{Minimize} \quad & J \\
\text{subject to} \quad 
& \sum_{c} x_c = K \\
& y_{v,s} \leq \sum_{c: u(v,c) \geq s} x_c \quad \forall v, s \\
& \sum_{v: u(v, c') \geq s} (1 - y_{v,s}) \leq J \quad \forall c', s \\
& x_c \in \{0, 1\} \quad \forall c \\
& y_{v,s} \in \{0,1\} \quad \forall v, s \\
& J \in \mathbb{Z}; J \geq 0
\end{align*}

This is a large integer program, with $O(N^2M)$ integer variables (since there can be up to $NM$ distinct utility levels). To obtain a more efficient program, we simply relax the $y$ variables to be fractional between 0 and 1, inclusive. At optimality, if a specific $y_{v,s}$ is strictly fractional, it can simply be set to 1 without violating any constraints (since $x$ is still required to be binary). This gives us a much more efficient integer program, with only $O(M)$ integer variables:

\subsubsection*{More Efficient Integer Program}
\begin{align*}
\text{Minimize} \quad & J \\
\text{subject to} \quad 
& \sum_{c} x_c = K \\
& y_{v,s} \leq \sum_{c: u(v,c) \geq s} x_c \quad \forall v, s \\
& \sum_{v: u(v, c') \geq s} (1 - y_{v,s}) \leq J \quad \forall c', s \\
& x_c \in \{0, 1\} \quad \forall c \\
& 0 \leq y_{v,s} \leq 1 \quad \forall v, s \\
& J \in \mathbb{Z}; J \geq 0
\end{align*}

Note that $J$ can also be similarly relaxed, but in practice we found that explicitly requiring $J$ to be integral actually allows commercial solvers to solve the problem faster, since the solver can stop when the difference between an upper bound and a lower bound is less than 1.


\end{document}